\documentclass[10pt,twocolumn,letterpaper]{article}

\usepackage{cvpr}
\usepackage{times}
\usepackage{epsfig}
\usepackage{graphicx}
\usepackage{amsmath}
\usepackage{amssymb}
\usepackage{color}
\usepackage{algorithm}
\usepackage{algorithmic}
\usepackage{multirow}

\usepackage[pagebackref=true,breaklinks=true,letterpaper=true,colorlinks,bookmarks=false]{hyperref}

\cvprfinalcopy 

\def\cvprPaperID{2557} 

\ifcvprfinal\fi
\begin{document}

\title{Deep Learning with Low Precision by Half-wave Gaussian Quantization}

\author{Zhaowei Cai\\
UC San Diego\\
{\tt\small zwcai@ucsd.edu}
\and
Xiaodong He\\
Microsoft Research Redmond\\
{\tt\small xiaohe@microsoft.com}
\and
Jian Sun\\
Megvii Inc.\\
{\tt\small sunjian@megvii.com}
\and
Nuno Vasconcelos\\
UC San Diego\\
{\tt\small nuno@ucsd.edu}
}

\maketitle

\begin{abstract}
The problem of quantizing the activations of a deep neural network is considered. An examination of the popular binary quantization approach shows that this consists of approximating a classical non-linearity, the hyperbolic tangent, by two functions: a piecewise constant $sign$ function, which is used in feedforward network computations, and a piecewise linear {\it hard tanh\/} function, used in the backpropagation step during network learning. The problem of approximating the ReLU non-linearity, widely used in the recent deep learning literature, is then considered. An half-wave Gaussian quantizer (HWGQ) is proposed for forward approximation and shown to have efficient implementation, by exploiting the statistics of of network activations and batch normalization operations commonly used in the literature. To overcome the problem of gradient mismatch, due to the use of
different forward and backward approximations, several piece-wise backward approximators are then investigated. The implementation of the resulting quantized network, denoted as HWGQ-Net, is shown to achieve much closer performance to full precision networks, such as AlexNet, ResNet, GoogLeNet and VGG-Net, than previously available low-precision networks, with 1-bit binary weights and 2-bit quantized activations.
\end{abstract}

\section{Introduction}

Deep neural networks have achieved state-of-the-art performance on computer
vision problems, such as classification \cite{DBLP:conf/nips/KrizhevskySH12,
DBLP:journals/corr/SimonyanZ14a,DBLP:conf/cvpr/SzegedyLJSRAEVR15,DBLP:conf/iccv/HeZRS15,DBLP:journals/corr/HeZRS15},
detection \cite{DBLP:conf/iccv/Girshick15,DBLP:conf/nips/RenHGS15,DBLP:conf/eccv/CaiFFV16}, etc. However, their complexity is an impediment to widespread
deployment in many applications of real world interest, where either
memory or computational resource is limited. This is due to two main issues:
large model sizes (50MB for GoogLeNet \cite{DBLP:conf/cvpr/SzegedyLJSRAEVR15}, 200M for ResNet-101 \cite{DBLP:journals/corr/HeZRS15}, 250MB for AlexNet \cite{DBLP:conf/nips/KrizhevskySH12}, or 500M for
VGG-Net \cite{DBLP:journals/corr/SimonyanZ14a}) and large computational
cost, typically requiring GPU-based implementations. This generated interest in compressed models with smaller memory footprints and computation.

Several works have addressed the reduction of model size, through
the use of quantization \cite{DBLP:conf/nips/CourbariauxBD15,DBLP:journals/corr/LinCMB15,DBLP:conf/icml/LinTA16},
low-rank matrix factorization \cite{DBLP:conf/bmvc/JaderbergVZ14,DBLP:conf/nips/DentonZBLF14},
pruning \cite{DBLP:conf/nips/HanPTD15,DBLP:journals/corr/HanMD15},
architecture design \cite{DBLP:journals/corr/LinCY13,DBLP:journals/corr/IandolaMAHDK16}, etc. Recently, it has been shown that weight compression by
quantization can achieve very large savings in memory, reducing each weight
to as little as 1 bit, with a marginal cost in classification accuracy \cite{DBLP:conf/nips/CourbariauxBD15,DBLP:journals/corr/LinCMB15}. However, it is less effective along the computational dimension, because the core network operation, implemented by each of its units, is the dot-product between a weight and an activation vector. On the other hand, complementing binary or quantized weights with quantized activations allows the replacement of expensive dot-products by logical and bit-counting operations. Hence, substantial speed ups are possible if, in addition to the weights, the inputs of each unit are binarized or quantized to low-bit.

It appears, however, that the quantization of activations is more
difficult than that of weights. For example,
\cite{DBLP:journals/corr/HubaraCSEB16,DBLP:conf/eccv/RastegariORF16} have
shown that, while it is possible to binarize weights with a marginal cost
in model accuracy, additional quantization of activations incurs nontrivial
losses for large-scale classification tasks, such as
object recognition on ImageNet \cite{DBLP:journals/ijcv/RussakovskyDSKS15}.
The difficulty is that binarization or quantization of activations
requires their processing with non-differentiable operators.
This creates problems for the gradient descent procedure, the
backpropagation algorithm, commonly used to learn deep networks.
This algorithm iterates between a feedforward step that
computes network outputs and a backpropagation step that computes the
gradients required for learning. The difficulty is that binarization or
quantization operators have step-wise responses that produce very weak
gradient signals during backpropagation, compromising learning efficiency.
So far, the problem has been addressed by using
continuous approximations of the operator used in the feedforward
step to implement the backpropagation step. This, however, creates a
mismatch between the model that implements the forward computations and
the derivatives used to learn it. In result, the model learned by the
backpropagation procedure tends to be sub-optimal.

In this work, we view the quantization operator, used in the feedforward
step, and the continuous approximation, used in the backpropagation step,
as two functions that approximate the activation function of each network
unit. We refer to these as the {\it forward\/} and {\it backward\/}
approximation of the activation function. We start by considering the
binary $\pm{1}$ quantizer, used
in \cite{DBLP:journals/corr/HubaraCSEB16,DBLP:conf/eccv/RastegariORF16},
for which these two functions can be seen as a discrete and a continuous
approximation of a non-linear activation function, the hyperbolic tangent,
frequently used in classical neural networks. This activation is, however,
not commonly used in recent deep learning literature, where the ReLU
nonlinearity \cite{DBLP:conf/icml/NairH10,DBLP:conf/icassp/ZeilerRMMYLNSVDH13,
DBLP:conf/iccv/HeZRS15} has achieved much greater preponderance.
This is exactly because it produces much stronger gradient magnitudes.
While the hyperbolic tangent or sigmoid non-linearities are squashing
non-linearities and mostly flat, the ReLU is an half-wave rectifier, of
linear response to positive inputs. Hence, while the derivatives
of the hyperbolic tangent are close to zero almost everywhere, the ReLU has
unit derivative along the entire positive range of the axis.

To improve the learning efficiency of quantized networks, we consider
the design of forward and backward approximation functions for the ReLU.
To discretize its linear component, we propose to use an optimal quantizer.
By exploiting the statistics of network activations and batch normalization
operations that are commonly used in the literature, we show that this can
be done with an half-wave Gaussian quantizer (HWGQ) that requires no
learning and is very efficient to compute. While some recent works have
attempted similar ideas \cite{DBLP:journals/corr/HubaraCSEB16,DBLP:conf/eccv/RastegariORF16},
their design of a quantizer is not sufficient to guarantee good deep learning performance. We address this problem by complementing this design with a study of suitable backward approximation functions that account for the mismatch between the forward model and the back propagated derivatives. This study suggests operations such as linearization, gradient clipping or gradient suppression for the implementation of the backward approximation. We show that a combination of the forward HWGQ with these backward operations produces very efficient low-precision networks, denoted as HWGQ-Net, with much closer performance to continuous models, such as AlexNet \cite{DBLP:conf/nips/KrizhevskySH12}, ResNet \cite{DBLP:journals/corr/HeZRS15}, GoogLeNet \cite{DBLP:conf/cvpr/SzegedyLJSRAEVR15} and VGG-Net \cite{DBLP:journals/corr/SimonyanZ14a}, than
other available low-precision networks in the literature. To the best of our knowledge, this is the first time that a single low-precision algorithm could achieve successes for so many popular networks. According to \cite{DBLP:conf/eccv/RastegariORF16}, theoretically HWGQ-Net (1-bit weights and 2-bit activations) has $\sim$32$\times$ memory and $\sim$32$\times$ convolutional computation savings. These suggest that the HWGQ-Net can be very useful for the deployment of state-of-the-art neural networks in real world applications.

\section{Related Work}

The reduction of model size is a popular goal in the deep learning
literature, due to its importance for the deployment of high performance neural
networks in real word applications. One strategy is to exploit the
widely known redundancy of neural network
weights \cite{DBLP:conf/nips/DenilSDRF13}. For example,
\cite{DBLP:conf/bmvc/JaderbergVZ14,DBLP:conf/nips/DentonZBLF14} proposed
low-rank matrix factorization as a way to decompose a large weight
matrix into several separable small matrices. This approach has been
shown most successful for fully connected layers. An alternative
procedure, known as connection
pruning \cite{DBLP:conf/nips/HanPTD15,DBLP:journals/corr/HanMD15},
consists of removing unimportant connections of a pre-trained model and
retraining. This has been shown to reduce the number of model parameters
by an order of magnitude without considerable loss in classification
accuracy. Another model compression strategy is to constrain the
model architecture itself, e.g. by removing fully connected layers,
using convolutional filters of small size, etc. Many state-of-the-art
deep networks, such as NIN \cite{DBLP:journals/corr/LinCY13},
GoogLeNet \cite{DBLP:conf/cvpr/SzegedyLJSRAEVR15} and
ResNet \cite{DBLP:journals/corr/HeZRS15}, rely on such design choices.
For example, SqueezeNet \cite{DBLP:journals/corr/IandolaMAHDK16} has been shown to achieve a parameter reduction
of $\sim$50 times, for accuracy comparable to that of AlexNet.
Moreover, hash functions have also been used to compress model
size \cite{DBLP:conf/icml/ChenWTWC15}.

Another branch of approaches for model compression is weight
binarization \cite{DBLP:conf/nips/CourbariauxBD15,DBLP:conf/eccv/RastegariORF16,DBLP:journals/corr/HubaraCSEB16} or quantization \cite{DBLP:journals/corr/LinCMB15,DBLP:conf/icml/LinTA16,DBLP:journals/corr/GongLYB14}. \cite{37631} used a
fixed-point representation to quantize weights of pre-trained neural
networks, so as to speed up testing on CPUs.
\cite{DBLP:journals/corr/GongLYB14} explored several alternative
quantization methods to decrease model size, showing that
procedures such as vector quantization, with $k$-means, enable 4$\sim$8 times
compression with minimal accuracy loss. \cite{DBLP:conf/icml/LinTA16}
proposed a method for fixed-point quantization based on the identification
of optimal bit-width allocations across network layers.
\cite{DBLP:journals/corr/LiL16,DBLP:journals/corr/LinCMB15} have shown
that ternary weight quantization into levels $\{-1, 0, 1\}$ can achieve
$16\times$ or $32\times$ model compression with slight accuracy loss, even
on large-scale classification tasks. Finally, \cite{DBLP:conf/nips/CourbariauxBD15} has shown that filter weights can be quantized to $\pm 1$ without
noticeable loss of classification accuracy on datasets such as CIFAR-10 \cite{krizhevsky2009learning}.

Beyond weight binarization, the quantization of activations has two additional benefits: 1) further speed-ups by replacement of the expensive
inner-products at the core of all network computations with logical and bit-counting operations; 2) training memory savings, by avoiding the large amounts of memory required to cache full-precision activations. Due to these, activation quantization has attracted some attention recently \cite{37631,DBLP:conf/icml/LinTA16,DBLP:journals/corr/HubaraCSEB16,DBLP:conf/eccv/RastegariORF16,DBLP:journals/corr/ZhouNZWWZ16,DBLP:journals/corr/LinT16a,DBLP:journals/corr/LinCMB15}. In \cite{37631}, activations were quantized with 8 bits, to
achieve speed-ups on CPUs. By performing the quantization after
network training, this work avoided the issues of nondifferentiable
optimization. \cite{DBLP:conf/icml/LinTA16} developed an optimal algorithm
for bit-width allocation across layers, but did not
propose a learning procedure for quantized neural networks.
Recently, \cite{DBLP:journals/corr/HubaraCSEB16,DBLP:conf/eccv/RastegariORF16,DBLP:journals/corr/ZhouNZWWZ16} tried to tackle the optimization
of networks with nondifferentiable quantization units,
by using a continuous approximation to the quantizer function in the
backpropagation step. \cite{DBLP:journals/corr/LinT16a} proposed
several potential solutions to the problem of gradient mismatch and
\cite{DBLP:journals/corr/LinCMB15,DBLP:journals/corr/ZhouNZWWZ16}
showed that gradients can be quantized with a small number of
bits during the backpropagation step. While some of these
methods have produced good results on datasets such as CIFAR-10,
none has produced low precision networks competitive with
full-precision models on large-scale classification tasks, such as
ImageNet \cite{DBLP:journals/ijcv/RussakovskyDSKS15}.

\section{Binary Networks}
\label{sec:binary networks}

We start with a brief review of the issues involved in the binarization of a deep network.

\subsection{Goals}

Deep neural networks are composed by layers of processing units
that roughly model the computations of the neurons found in the mammalian
brain. Each unit computes an activation function of the form
\begin{equation}
  \label{equ:activation}
  z=g(\textbf{w}^T\textbf{x}),
\end{equation}
where $\textbf{w}\in\mathbb{R}^{c \cdot w \cdot h}$ is a weight vector,
$\textbf{x}\in\mathbb{R}^{c \cdot w \cdot h}$ an input vector,
and $g(\cdot)$ a non-linear function. 
A convolutional network implements layers of these units, where
weights are usually represented as a tensor
$\textbf{W}\in\mathbb{R}^{c \times w \times h}$. The dimensions $c$, $w$ and $h$
are defined by the number of filter channels, width and height, respectively. Since this basic computation
is repeated throughout the network and modern networks have very large
numbers of units, the structure of~(\ref{equ:activation}) is the main
factor in the complexity of the overall model. This complexity can be
a problem for applications along two dimensions. The first is the
large memory footprint required to store weights $\textbf{w}$. The second
is the computational complexity required to compute large numbers of
dot-products $\textbf{w}^T\textbf{x}$. Both difficulties are compounded by
the requirement of floating point storage of weights and floating point
arithmetic to compute dot-products, which are not practical for many
applications. This has motivated interest in low-precision networks \cite{DBLP:journals/corr/HubaraCSEB16,DBLP:conf/eccv/RastegariORF16,DBLP:journals/corr/ZhouNZWWZ16}.

\subsection{Weight Binarization}

An effective strategy to binarize the weights $\textbf{W}$ of convolutional filters, which we adopt
in this work, has been proposed by \cite{DBLP:conf/eccv/RastegariORF16}.
This consists of approximating the full precision weight
matrix $\textbf{W}$, used to compute the activations of~(\ref{equ:activation}) for all the units, by the product of a binary matrix $\textbf{B}\in\{+1,-1\}^{c\times{w}\times{h}}$
and a scaling factor $\alpha\in\mathbb{R}^+$, such that
$\textbf{W}\approx\alpha\textbf{B}$. A convolutional operation on
input $\textbf{I}$ can then be approximated by
\begin{equation}
  \textbf{I}\ast\textbf{W}\approx\alpha(\textbf{I}\oplus\textbf{B}),
  \label{eq:wapprox}
\end{equation}
where $\oplus$ denotes a multiplication-free
convolution.~\cite{DBLP:conf/eccv/RastegariORF16}
has shown that an optimal approximation can be achieved with $\textbf{B}^*=sign(\textbf{W})$ and $\alpha^*=\frac{1}{cwh}\|\textbf{W}\|_1$.
More details can be found in \cite{DBLP:conf/eccv/RastegariORF16}.

While binary weights tremendously reduce the memory footprint of the model,
they do not fully solve the problem of computational complexity.
Since $\textbf{I}$ consists of either the activations of a previous layer or some transformation of the image to classify, it is usually represented with full precision. Hence,~(\ref{eq:wapprox}) requires floating point
arithmetic and produces a floating point result. Substantial further
reductions of complexity can be obtained by the binarization of $\textbf{I}$,
which enables the implementation of dot products with logical and bit-counting
operations \cite{DBLP:journals/corr/HubaraCSEB16,DBLP:conf/eccv/RastegariORF16}.

\begin{figure}[!t]
\begin{minipage}[b]{.49\linewidth}
\centering
\centerline{\epsfig{figure=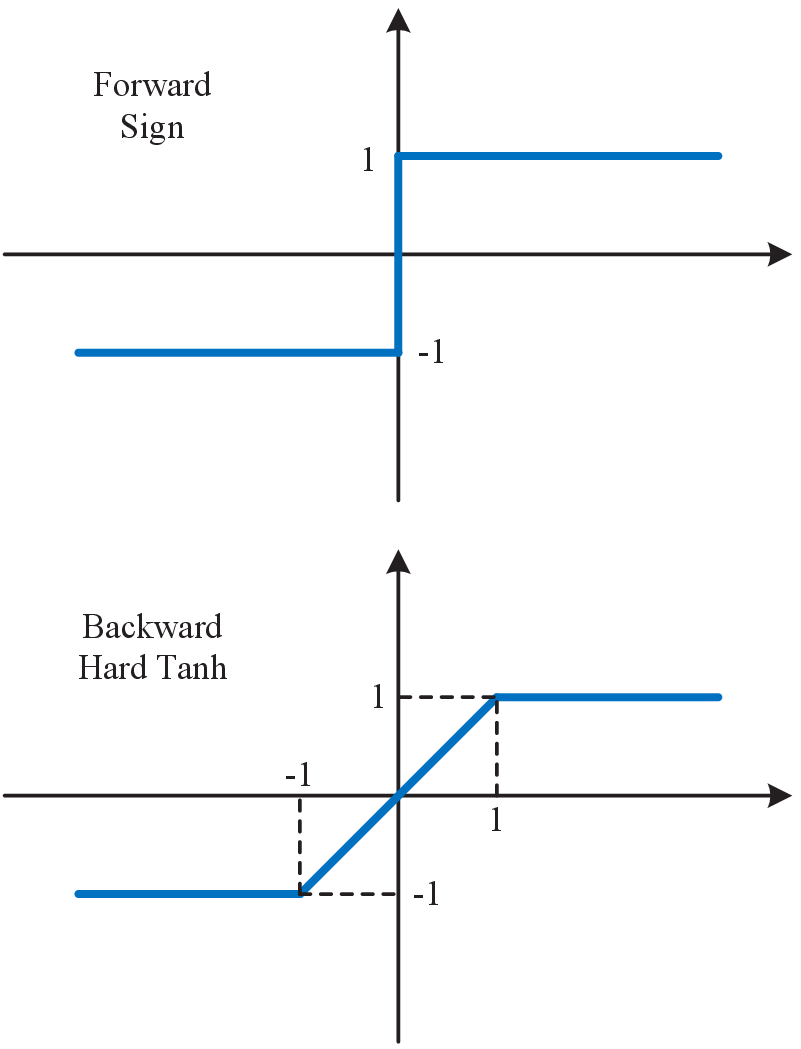,width=3.5cm,height=4.4cm}}
\end{minipage}
\hfill
\begin{minipage}[b]{.49\linewidth}
\centering
\centerline{\epsfig{figure=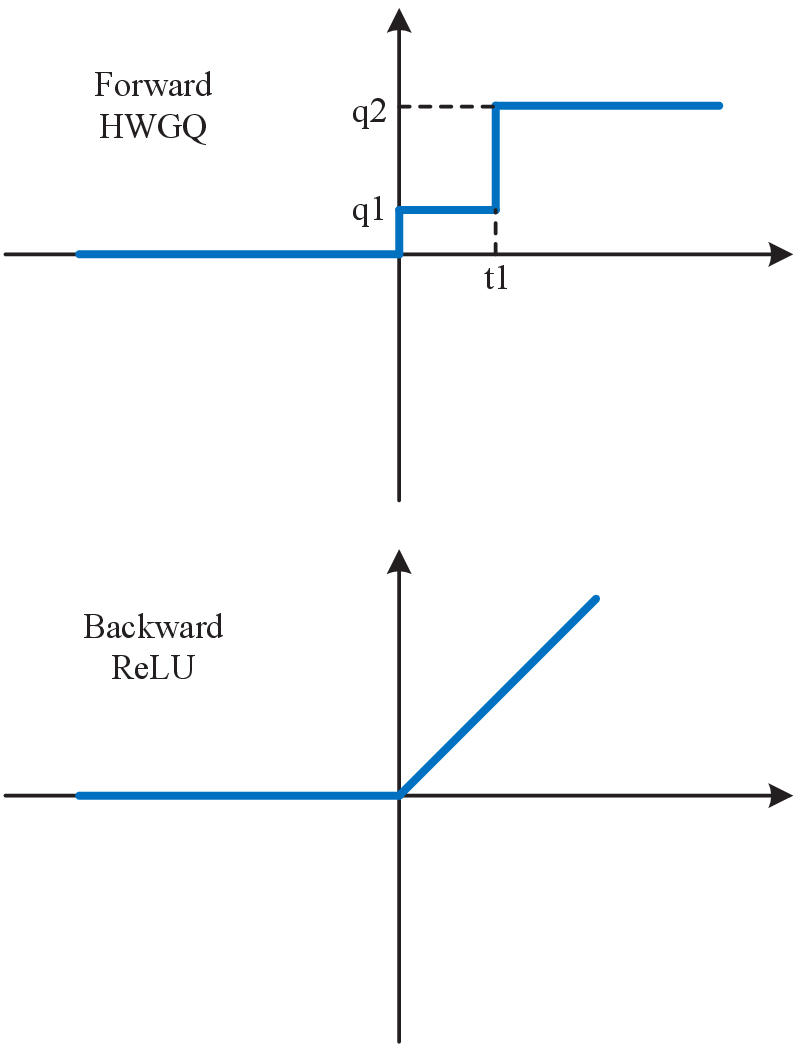,width=3.5cm,height=4.4cm}}
\end{minipage}
\caption{Forward and backward functions for binary $sign$ (left) and half-wave Gaussian quantization (right) activations.}
\label{fig:activation}
\end{figure}

\subsection{Binary Activation Quantization}
\label{subsec:binary quantization}

This problem has been studied in the literature, where the use of binary
activations has recently become
popular \cite{DBLP:conf/eccv/RastegariORF16,DBLP:journals/corr/HubaraCSEB16,DBLP:journals/corr/ZhouNZWWZ16}.
This is usually implemented by replacing $g(x)$ in~(\ref{equ:activation})
with the $sign$ non-linearity
\begin{equation}
  \label{equ:activation sign}
  z=sign(x)=\left\{
    \begin{array}{cl}+1, &\quad\textrm{if $x\geq{0}$,}\\
      -1, &\quad\textrm{otherwise}
    \end{array}\right.
\end{equation}
shown in Figure \ref{fig:activation}. \cite{DBLP:conf/eccv/RastegariORF16}
has also considered rescaling the binarized outputs $z$ by a
factor $\beta$, but found this to be unnecessary.

\begin{figure*}[!t]
\begin{minipage}[b]{.15\linewidth}
\centering
\centerline{\epsfig{figure=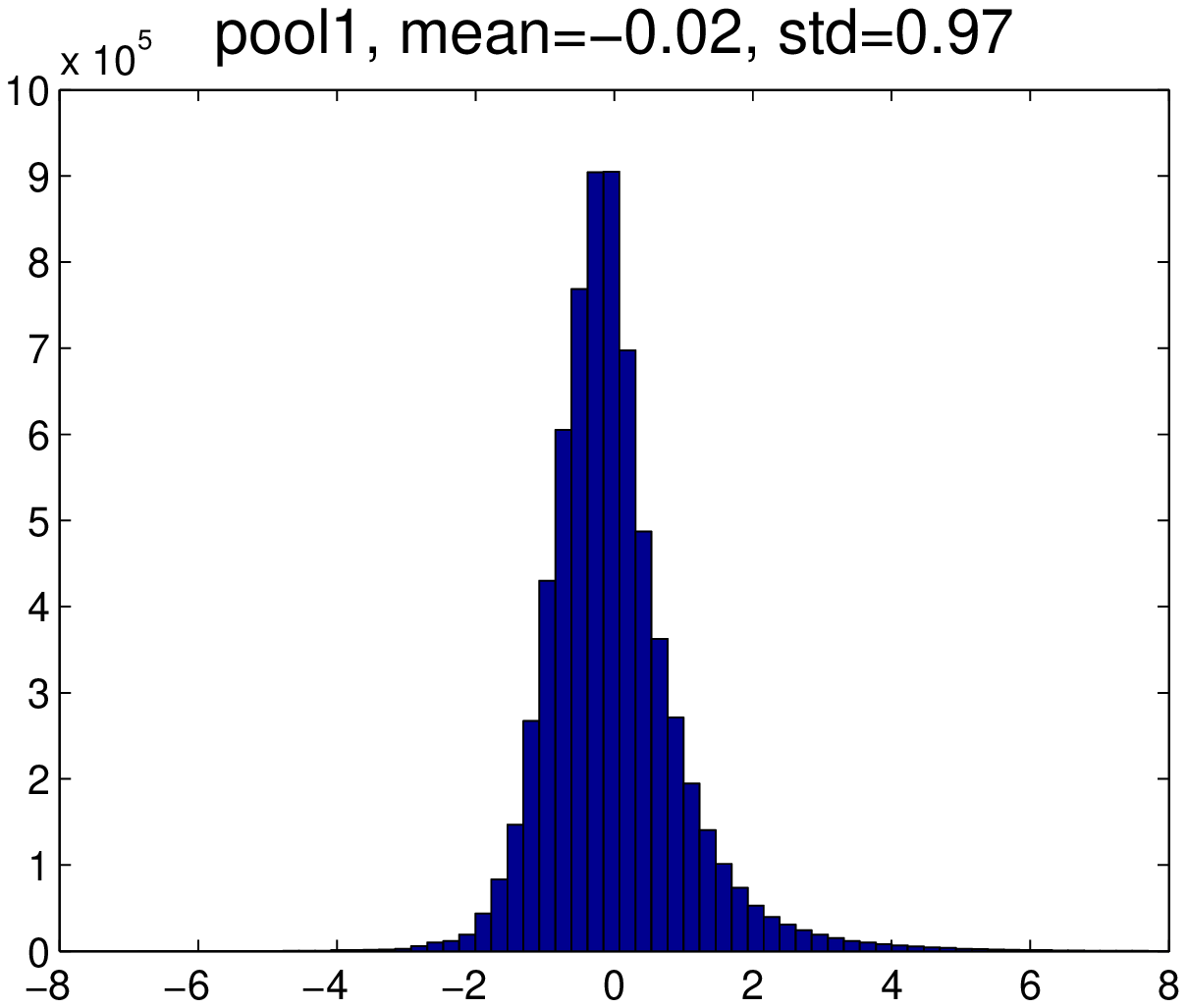,width=3.2cm,height=2.4cm}}
\end{minipage}
\hfill
\begin{minipage}[b]{.15\linewidth}
\centering
\centerline{\epsfig{figure=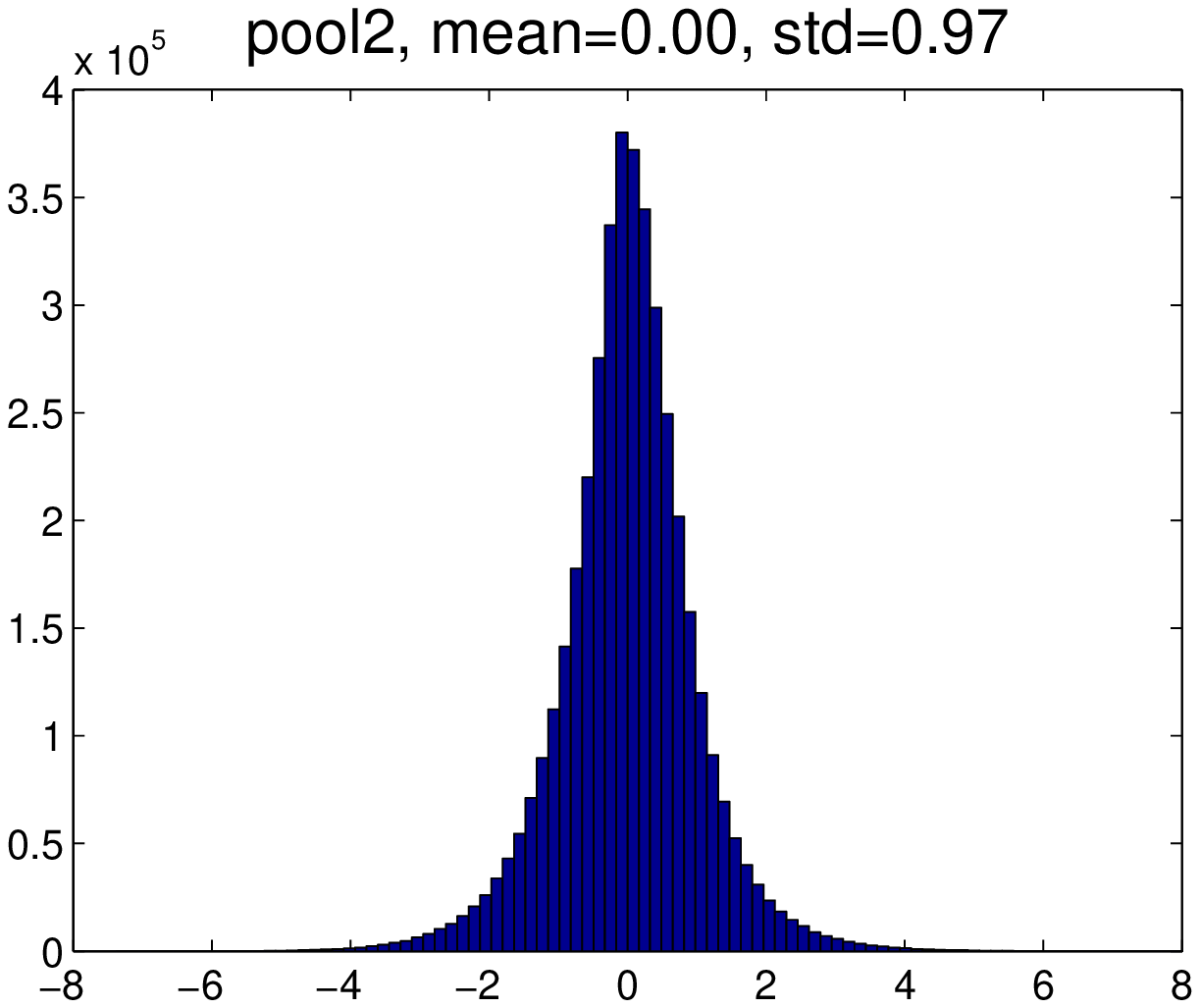,width=3.2cm,height=2.4cm}}
\end{minipage}
\hfill
\begin{minipage}[b]{.15\linewidth}
\centering
\centerline{\epsfig{figure=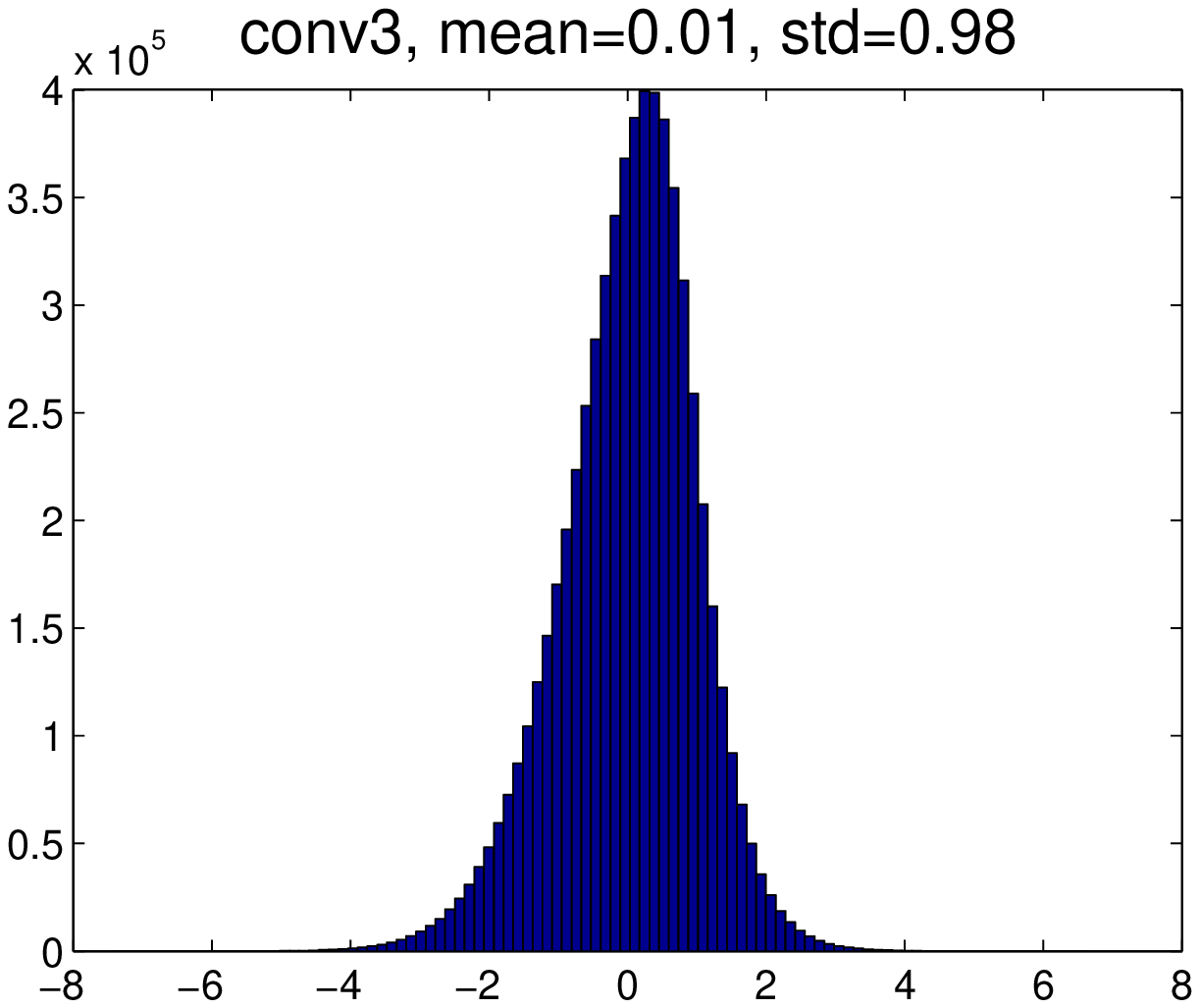,width=3.2cm,height=2.4cm}}
\end{minipage}
\hfill
\begin{minipage}[b]{.15\linewidth}
\centering
\centerline{\epsfig{figure=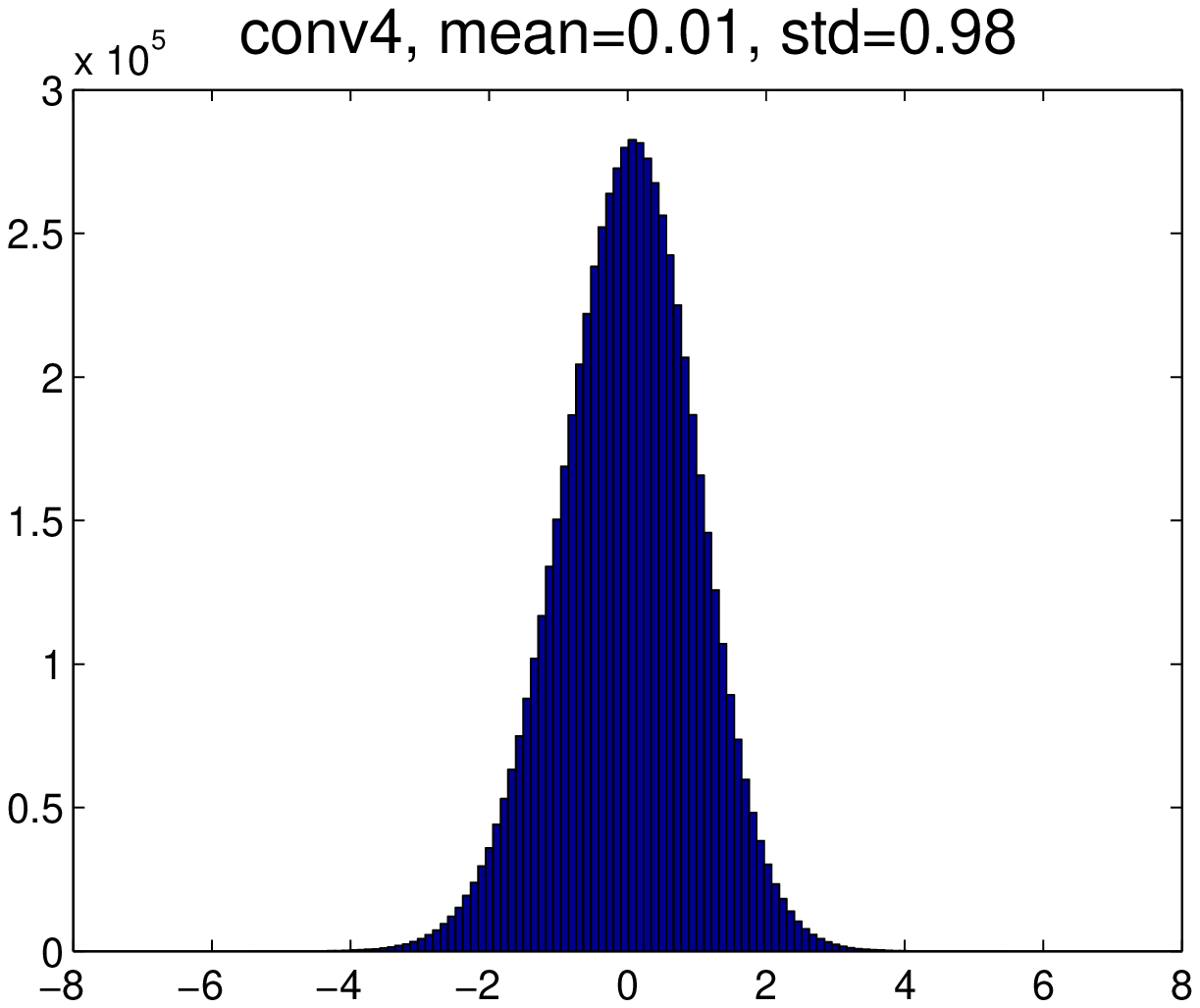,width=3.2cm,height=2.4cm}}
\end{minipage}
\hfill
\begin{minipage}[b]{.15\linewidth}
\centering
\centerline{\epsfig{figure=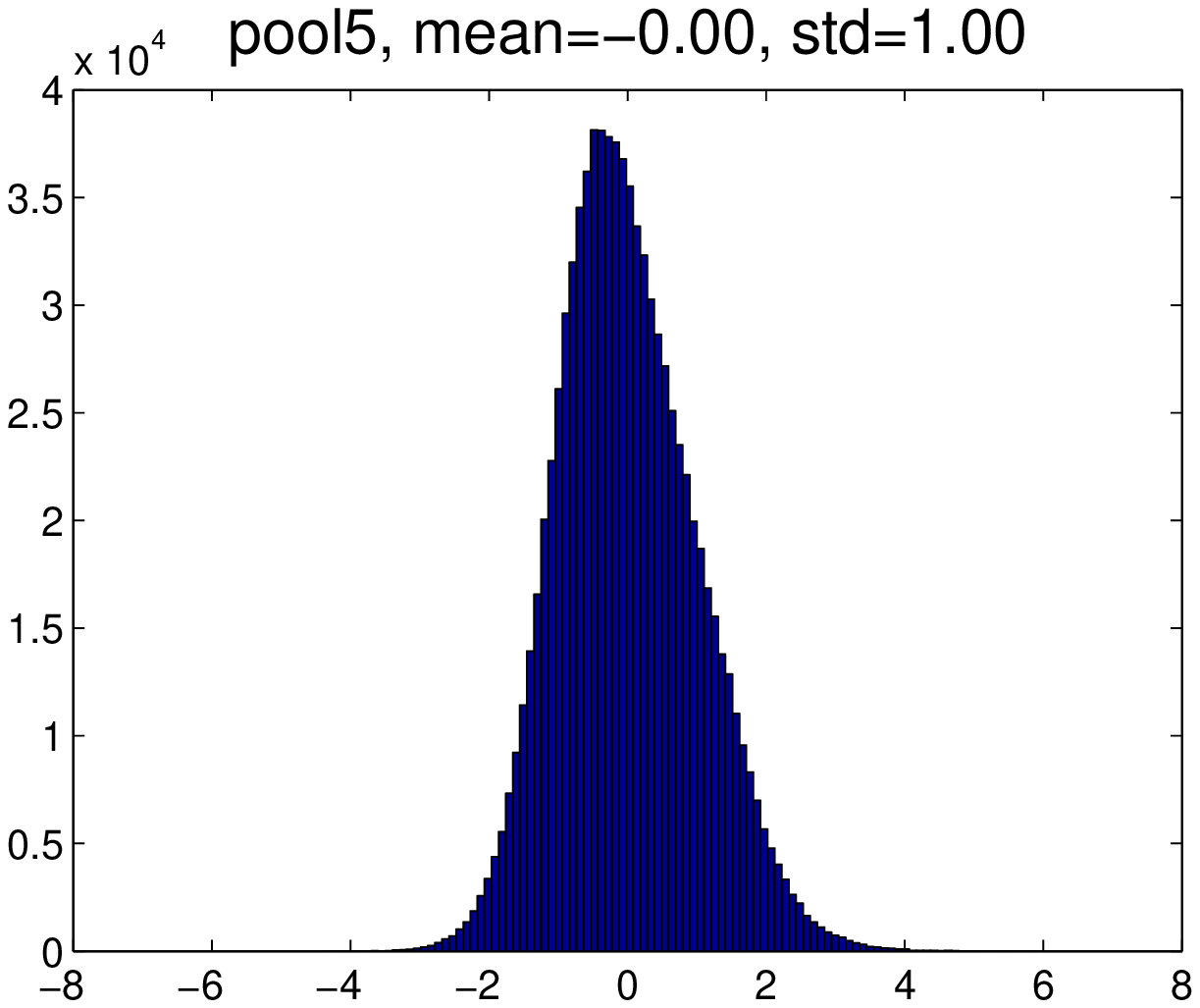,width=3.2cm,height=2.4cm}}
\end{minipage}
\hfill
\begin{minipage}[b]{.15\linewidth}
\centering
\centerline{\epsfig{figure=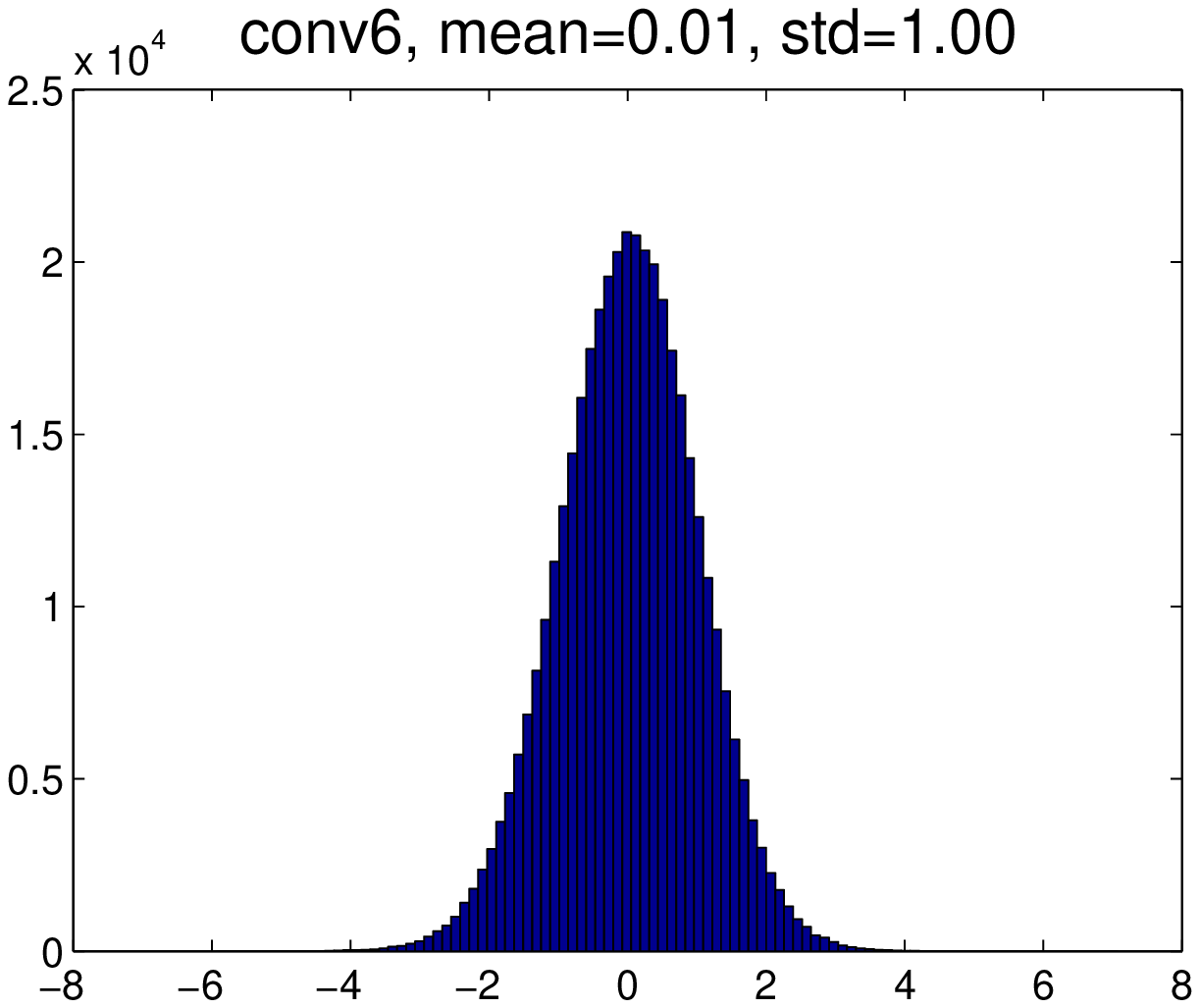,width=3.2cm,height=2.4cm}}
\end{minipage}
\caption{Dot-product distributions on different layers of AlexNet with
  binary weights and quantized activations (100 random images).}
\label{fig:distribution}
\end{figure*}

While the adoption of~(\ref{equ:activation sign}) greatly simplifies the
feedforward computations of the deep model, it severely increases the
difficulty of learning. This follows from the fact that the derivative of the sign function is zero almost everywhere. Neural networks are learned
by minimizing a cost $C$ with respect to the weights $\bf w$.
This is done with the backpropagation algorithm, which decomposes
these derivatives into a series of simple computations. Consider
the unit of~(\ref{equ:activation}). The derivative of $C$ with respect to
$\bf w$ is
\begin{equation}
  \label{equ:gradient mismatch}
  \frac{\partial{C}}{\partial{\bf w}}=
  \frac{\partial{C}}{\partial{z}} g^\prime({\bf w}^T{\bf x}) {\bf x}.
\end{equation}
These derivatives are computed for all units during the backpropagation step. When $g(x)$ is replaced by~(\ref{equ:activation sign}), the derivative $g^\prime({\bf w}^T{\bf x})$ is zero almost everywhere
and the gradient magnitudes tend to be very small. In result, the
gradient descent algorithm does not converge to minima of the cost.
To overcome this problem, \cite{DBLP:journals/corr/HubaraCSEB16}
proposed to use an alternative function, {\it hard tanh\/}, which we denote
by $\widetilde{sign}$, in the backpropagation step. This function is
shown in Figure \ref{fig:activation}, and has derivative
\begin{equation}
  \label{equ:sign derivative}
  \widetilde{sign}'(x)=\left\{
    \begin{array}{cl}1, &\quad\textrm{if $|x|\leq{1}$}\\
      0, &\quad\textrm{otherwise.}
    \end{array}\right.
\end{equation}

In this work, we denote~(\ref{equ:activation sign}) as the forward
and~(\ref{equ:sign derivative}) as the backward approximations of the
activation non-linearity $g(x)$ of~(\ref{equ:activation}). These
approximations have two main problems. The first is that they
approximate the hyperbolic tangent
\begin{equation}
  g(x) = tanh(x) = \frac{e^x - e^{-x}}{e^x + e^{-x}}.\nonumber
\end{equation}
This (and the closely related sigmoid function) is a squashing non-linearity that replicates the saturating behavior of neural firing rates. For this
reason, it was widely used in the classical neural net literature. However,
squashing non-linearities have been  close to abandoned in recent deep learning literature, because the saturating behavior emphasizes the problem of vanishing derivatives, compromising the effectiveness of backpropagation. The second problem is that the discrepancy between the approximation of
$g(x)$ by the forward $sign$ and by the backward $\widetilde{sign}$ creates a mismatch between the feedforward model and the derivatives used to learn it. In result, backpropagation can be highly suboptimal. This is called the ``gradient mismatch'' problem \cite{DBLP:journals/corr/LinT16a}.

\section{Half-wave Gaussian Quantization}
\label{sec:hwgq}

In this section, we propose an alternative quantization strategy, which
is based on the approximation of the ReLU non-linearity.

\subsection{ReLU}

The ReLU is the half-wave rectifier defined by \cite{DBLP:conf/icml/NairH10}
\begin{equation}
  g(x) = \max(0, x).
  \label{equ:relu}
\end{equation}
It is now well known that, when compared to squashing non-linearities,
its use in~(\ref{equ:activation}) significantly improves the efficiency of the backpropagation algorithm. It thus seems more sensible
to rely on ReLU approximations for network quantization than those of the
previous section. We propose a quantizer $Q(x)$ to approximate~(\ref{equ:relu}) in the feedforward step and a suitable piecewise linear approximation
$\widetilde{Q}(x)$ for the backpropagation step.

\subsection{Forward Approximation}
\label{subsec:gaussian}

A quantizer is a piecewise constant function
\begin{equation}
\label{equ:gaussian quant}
Q(x)=q_i, \quad if \quad x\in(t_i,t_{i+1}],
\end{equation}
that maps all values of $x$ within quantization interval $(t_i,t_{i+1}]$
into a quantization level $q_i\in\mathbb{R}$, for $i=1,\cdots,m$. In
general, $t_1=-\infty$ and $t_{m+1}=\infty$. This generalizes the $sign$
function, which can be seen as a $1$-bit quantizer. A quantizer is
denoted uniform if
\begin{equation}
  \label{equ:unif quant}
  q_{i+1}-q_i = \Delta , \quad \forall i,
\end{equation}
where $\Delta$ is a constant quantization step. The quantization levels $q_i$ act as the reconstruction values for $x$, under the constraint of reduced precision. Since, for any $x$, it suffices to store the quantization index $i$ of~(\ref{equ:gaussian quant}) to recover the quantization level $q_i$, non-uniform quantizer requires $\log_{2} m$ bits of storage per activation $x$. However, it usually requires more than $\log_{2} m$ bits to represent $x$ during arithmetic operations, in which it is $q_i$ to be used instead of index $i$. For a uniform quantizer, where $\Delta$ is a universal scaling factor that can be placed in evidence, it is intuitive to store any $x$ by $\log_{2} m$ bits without storing the indexes. The same holds for arithmetic computation.

Optimal quantizers are usually defined in the mean-squared error sense, i.e.
\begin{align}
\label{eq:optq}
  Q^*(x) &=\arg\min_Q E_x[(Q(x)-x)^2] \\\nonumber
         &=\arg\min_Q \int p(x) (Q(x)-x)^2 dx
\end{align}
where $p(x)$ is the probability density function of $x$. Hence, the optimal
quantizer of the dot-products of~(\ref{equ:activation}) depends on their
statistics. While the optimal solution $Q^*(x)$ of (\ref{eq:optq}) is usually non-uniform, a uniform solution $Q^*(x)$ is available by adding the uniform constraint of (\ref{equ:unif quant}) to (\ref{eq:optq}). Given dot product samples, the optimal solution of~(\ref{eq:optq}) can be obtained by Lloyd's algorithm \cite{DBLP:journals/tit/Lloyd82}, which is similar to $k$-means algorithm. This, however, is an iterative algorithm. Since a different quantizer must be designed per network unit, and this quantizer changes with the backpropagation iteration, the straightforward application of this procedure is computationally intractable.

This difficulty can be avoided by exploiting the statistical
structure of the activations of deep networks. For example,
\cite{DBLP:journals/nn/HyvarinenO00,DBLP:conf/icml/IoffeS15} have noted that the dot-products of (\ref{equ:activation}) tend to have a
symmetric, non-sparse distribution, that is close to Gaussian. Taking
into account the fact that ReLU is a half-wave rectifier,
this suggests the use of the half-wave Gaussian quantizer (HWGQ),
\begin{equation}
\label{equ:half wave}
Q(x)=\left\{
  \begin{array}{cl}q_i, &\quad\textrm{if $x\in(t_i,t_{i+1}]$,}\\
    0, &\quad\textrm{$x\leq{0}$,}
  \end{array}\right.
\end{equation}
where $q_i\in\mathbb{R}^+$ for $i=1,\cdots,m$ and $t_i\in\mathbb{R}^+$ for
$i=1,\cdots,m+1$ ($t_1=0$ and $t_{m+1}=\infty$) are the optimal
quantization parameters for the Gaussian distribution. The adoption
of the HWGQ guarantees that these parameters only depend on the mean and
variance of the dot-product distribution. However, because these can vary
across units, it does not eliminate the need for the repeated application
of Lloyd's algorithm across the network.

This problem can be alleviated by resorting to batch
normalization \cite{DBLP:conf/icml/IoffeS15}. This is a widely used
normalization technique, which forces the responses of each network
layer to have zero mean and unit variance. We apply this normalization to the dot-products, with the result illustrated
in Figure \ref{fig:distribution}, for a number of AlexNet units of
different layers. Although the distributions are
not perfectly Gaussian and there are minor differences between them, they
are all close to Gaussian with zero mean and unit variance. It follows that the optimal quantization parameters $q_i^*$ and $t_i^*$ are
approximately identical across units, layers and
backpropagation iterations. Hence, Lloyd's algorithm can be applied once, with data from the entire network. In fact, because all distributions
are approximately Gaussian of zero mean and unit variance, the
quantizer can even be designed from samples of this distribution.
In our implementation, we drew $10^6$ samples from a standard Gaussian distribution of zero mean and unit variance, and obtained the optimal quantization parameters by Lloyd's algorithm. The resulting parameters $t_i^*$ and $q_i^*$ were used to parametrize a single HWGQ that was used in all layers, after batch normalization of dot-products.

\subsection{Backward Approximation}
\label{subsec:backward relus}

Since the HWGQ is a step-wise constant function, it has zero derivative
almost everywhere. Hence, the approximation of $g(x)$ by
$Q(x)$ in~(\ref{equ:gradient mismatch}) leads to the problem of vanishing
derivatives. As in Section~\ref{sec:binary networks}, a piecewise linear
function $\widetilde{Q}(x)$ can be used during the backpropagation step to
avoid weak convergence. In summary, we seek a piece-wise function
that provides a good approximation to the ReLU and to the HWGQ.
We next consider three possibilities.

\subsubsection{Vanilla ReLU}
\label{subsubsec:vanilla relu}

Since the ReLU of~(\ref{equ:relu}) is already a piece-wise linear
function, it seems sensible to use the ReLU itself, denoted the {\it vanilla ReLU\/}, as the backward
approximation function. This corresponds to using the derivative
\begin{equation}
  \label{equ:gaussian derivative}
  \widetilde{Q}'(x)=\left\{
    \begin{array}{cl}1, &\quad\textrm{if $x>0$,}\\
      0, &\quad\textrm{otherwise}
    \end{array}\right.
\end{equation}
in~(\ref{equ:gradient mismatch}). The forward and backward
approximations $Q(x)$ and $\widetilde{Q}(x)$ of the ReLU are
shown in Figure \ref{fig:activation}. Note that, while the backward
approximation is exact, it is not equal to the forward approximation. Hence, there is a gradient mismatch. This mismatch is particularly large for large
values of $x$. Note that, for $x > 0$, the approximation of $Q(x)$ by the
ReLU has error $|Q(x)-x|$. This is usually upper bounded
by $(t_{i+1}-q_i)$ for $x\in(t_i,t_{i+1}]$ but unbounded
when $x \in (t_m,\infty)$. Since these are the values on the tail of
the distribution of $x$, the ReLU is said to have a large mismatch
with $Q(x)$ ``on the tail.'' Due to this, when the ReLU is used to
approximate $g'(x)$ in~(\ref{equ:gradient mismatch}), it can originate
very inaccurate gradients for dot-products on the tail of the
dot-product distribution. In our experience, these inaccurate gradients
can make the learning algorithm unstable.

This is a classical problem in the robust estimation literature, where
outliers can unduly influence the performance of a learning
algorithm~\cite{huber1964robust}. For quantization, where $Q(x)$ assumes
that values of $x$ beyond $q_m$ have very low probability, large dot-products are effectively outliers. The classical solution to mitigate the influence of these outliers is to limit the growth rate of the error function. Since this is $|Q(x)-x|$, the problem is the
monotonicity of the ReLU beyond $x = q_m$. To address it,
we investigated alternative backwards approximation functions of slower
growth rate.

\begin{figure}[!t]
\centering
\centerline{\epsfig{figure=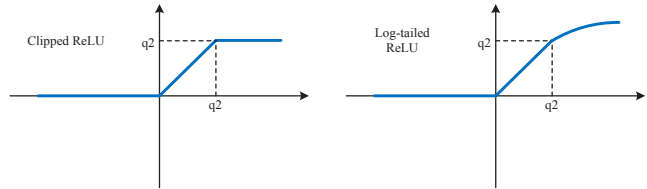,width=8.5cm,height=2.5cm}}
\caption{Backward piece-wise activation functions of clipped ReLU and log-tailed ReLU.}
\label{fig:relus}
\end{figure}

\subsubsection{Clipped ReLU}
\label{subsubsec:clipped relu}

The first approximation, denoted the {\it clipped ReLU\/}, is
identical to the vanilla ReLU in $(-\infty, q_m]$ but constant beyond $x = q_m$,
\begin{equation}
  \label{equ:clipped relu}
\widetilde{Q}_{c}(x)=\left\{
  \begin{array}{cl}q_m, &\quad x > q_m,\\
    x, &\quad x\in(0,q_m],\\
    0, &\quad\textrm{otherwise.}
\end{array}\right.
\end{equation}
Its use to approximate $g'({\bf w}^T{\bf x})$ in~(\ref{equ:gradient mismatch})
guarantees that there is no mismatch on the tail. Gradients are non-zero
only for dot-products in the interval $(0,q_m]$. We refer to this as \textit{ReLU clipping}. As illustrated in Figure \ref{fig:relus}, the clipped ReLU is a better match for the HWGQ than the vanilla ReLU. This idea is similar to the use of gradient clipping in \cite{DBLP:conf/icml/PascanuMB13}: clipping gradients to enable stable optimization, especially for very deep networks, e.g. recurrent neural network. In our experiments, ReLU clipping is also proved very useful to guarantee stable neural network optimization.

\subsubsection{Log-tailed ReLU}
\label{subsubsec:logtail relu}

Ideally, a network with quantized activations should approach the performance of
a full-precision network as the number of quantization levels $m$ increases. The sensitivity of the vanilla ReLU
approximation to outliers limits the performance of low precision networks
(low $m$). While the clipped ReLU alleviates this problem, it can impair
network performance due to
the loss of information in the clipped interval $(q_m,\infty)$.
An intermediate solution is to use, in this interval, a function
whose growth rate is in between that of the clipped ReLU (zero derivative)
and the ReLU (unit derivative). One possibility is to enforce logarithmic
growth on the tail, according to
\begin{equation}
  \label{equ:log relu}
  \widetilde{Q}_{l}(x)=\left\{
    \begin{array}{cl} q_m + \log(x-\tau), &\quad x>q_m,\\
      x, &\quad x\in(0,q_m],\\
      0, &\quad x\leq{0},
    \end{array}\right.
\end{equation}
where $\tau=q_m-1$. This is denoted the {\it log-tailed ReLU\/} and is compared to the ReLU and
clipped ReLU in Figure~\ref{fig:relus}. It has derivative
\begin{equation}
  \label{equ:log relu derivative}
  \widetilde{Q}_{l}^{'}(x)=\left\{
    \begin{array}{cl}
      1/(x-\tau), &\quad x>q_m,\\
      1, &\quad x\in(0,q_m],\\
      0, &\quad x\leq{0}.
    \end{array}\right.
\end{equation}
When used to approximate $g'(x)$ in~(\ref{equ:gradient mismatch}),
the log-tailed ReLU is identical to the vanilla ReLU for dot products of
amplitude smaller than $q_m$, but gives decreasing weight to
amplitudes larger than this. It behaves like the vanilla ReLU (unit derivative) for $x \approx q_m$ but converges to the clipped ReLU (zero derivative) as $x$ grows to infinity.

\section{Experimental Results}

The proposed HWGQ-Net was evaluated on ImageNet (ILSVRC2012)
\cite{DBLP:journals/ijcv/RussakovskyDSKS15}, which has $\sim$1.2M
training images from 1K categories and 50K validation images. The evaluation
metrics were top-1 and top-5 classification accuracy. Several popular
networks were tested: AlexNet \cite{DBLP:conf/nips/KrizhevskySH12},
ResNet \cite{DBLP:journals/corr/HeZRS15}, a variant of
VGG-Net \cite{DBLP:journals/corr/SimonyanZ14a,DBLP:conf/iccv/HeZRS15}, and
GoogLeNet \cite{DBLP:conf/cvpr/SzegedyLJSRAEVR15}. Our implementation is
based on Caffe \cite{DBLP:conf/mm/JiaSDKLGGD14}, and the source code is
available at https://github.com/zhaoweicai/hwgq.

\subsection{Implementation Details}
\label{subsec:details}

In all ImageNet experiments, training images were resized to 256$\times$256,
and a 224$\times$224 (227$\times$227 for AlexNet) crop was randomly sampled
from an image or its horizontal flip. Batch normalization
\cite{DBLP:conf/icml/IoffeS15} was applied before each quantization layer,
as discussed in Section \ref{subsec:gaussian}. Since weight binarization provides regularization constraints, the ratio of dropout \cite{DBLP:journals/corr/abs-1207-0580} was set as 0.1 for networks with binary weights and full activations, but no dropout was used for networks
with quantized activations regardless of weight precision. All networks were learned from scratch. No data augmentation was used other than standard random image flipping and cropping. SGD was used for parameter learning. No bias term was used for binarized weights. Similarly to \cite{DBLP:conf/eccv/RastegariORF16}, networks with quantized activations used max-pooling before batch normalization, which is denoted ``layer re-ordering''. As in \cite{DBLP:conf/eccv/RastegariORF16,DBLP:journals/corr/ZhouNZWWZ16}, the first and last network layers had full precision. Evaluation was based solely on central 224$\times$224 crop.

On AlexNet \cite{DBLP:conf/nips/KrizhevskySH12} experiments, the mini-batch
size was 256, weight decay 0.0005, and learning rate started at 0.01. For
ResNet, the parameters were the same as in \cite{DBLP:journals/corr/HeZRS15}.
For the variant of VGG-Net, denoted VGG-Variant, a smaller version of
model-A in \cite{DBLP:conf/iccv/HeZRS15}, only 3 convolutional layers were
used for input size of 56, 28 and 14, and the ``spp'' layer was removed.
The mini-batch size was 128, and learning rate started at 0.01. For
GoogLeNet \cite{DBLP:conf/cvpr/SzegedyLJSRAEVR15}, the branches for side losses
were removed, in the inception layers, max-pooling was removed and the channel
number of the ``reduce'' 1$\times$1 convolutional layers was increased to
that of their following 3$\times$3 and 5$\times$5 convolutional layers.
Weight decay was 0.0002 and the learning strategy was similar to
ResNet \cite{DBLP:journals/corr/HeZRS15}. For all networks tested, momentum
was 0.9, batch normalization was used, and when mini-batch size was 256 (128),
the learning rate was divided by 10 after every 50K (100K) iterations, 160K
(320K) in total. Only AlexNet, ResNet-18 and VGG-Variant were explored in
the following ablation studies. In all tables and figures, ``FW'' indicates
full-precision weights, ``BW'' binary weights, and ``Full'' full-precision weights and activations.

\begin{table}[t]
\centering \scriptsize \setlength{\tabcolsep}{5.0pt}
\vspace{0.1cm} \caption{Full-precision Activation Comparison for AlexNet.}
\label{tab:activation comp}
\begin{tabular}
{c||c|c|c|c|c}\hline
&Full &FW+$\widetilde{sign}$ &FW+$\widetilde{Q}$ &BW+$\widetilde{sign}$ &BW+$\widetilde{Q}$ \\\hline
Top-1    &55.7 &46.7  &55.7 &43.9  &53.9\\
Top-5    &79.3 &71.0  &79.3 &68.3  &77.3\\\hline
\end{tabular}
\end{table}

\subsection{Full-precision Activation Comparison}

Before considering the performance of the forward quantized activation functions $sign(x)$ and $Q(x)$, we compared the performance of the continuous $\widetilde{sign}(x)$ (hard tanh) and $\widetilde{Q}(x)$ (ReLU) as activation function. In this case, there is no activation quantization nor forward/backward gradient mismatch. AlexNet results
are presented in Table \ref{tab:activation comp},
using identical settings for $\widetilde{sign}(x)$ and $\widetilde{Q}(x)$,
for fair comparison. As expected from the discussion of
Sections \ref{sec:binary networks} and \ref{sec:hwgq},
$\widetilde{Q}(x)$ achieved substantially better performance
than $\widetilde{sign}(x)$, for both FW and BW networks.
The fact that these results upper bound the performance achievable when
quantization is included suggests that $sign(x)$ is not a good choice for
quantization function. $Q(x)$, on the other hand, has a fairly reasonable
upper bound.

\begin{table}[t]
\centering \scriptsize \setlength{\tabcolsep}{6.0pt}
\vspace{0.1cm} \caption{Low-bit Activation Comparison.}
\label{tab:low precision}
\begin{tabular}
{c|c||c|c|c|c|c}\hline
\multicolumn{2}{c||}{Model} &Full &BW &FW+$Q$ &BW+$sign$ &BW+$Q$\\\hline\hline
\multirow{2}{*}{AlexNet} &Top-1 &55.7 &52.4 &49.5 &39.5 &46.8\\
&Top-5 &79.3 &75.9 &73.7 &63.6 &71.0\\\hline
\multirow{2}{*}{ResNet-18} &Top-1 &66.3 &61.3 &37.5 &42.1 &33.0\\
&Top-5 &87.5 &83.6 &61.9 &67.1 &56.9\\\hline
\multirow{2}{*}{VGG-Variant} &Top-1  &68.6 &65.5 &48.3 &50.1 &44.1\\
&Top-5 &88.9 &86.5 &72.3 &74.3 &68.7\\\hline
\end{tabular}
\end{table}

\begin{figure}[!t]
\centering
\centerline{\epsfig{figure=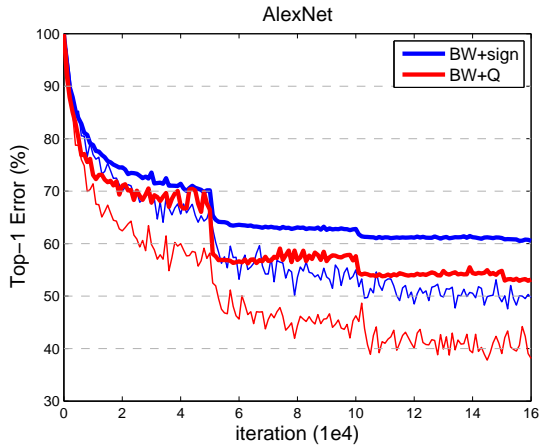,width=8cm,height=6cm}}
\caption{The error curves of training (thin) and test (thick) for $sign(x)$ and $Q(x)$ (HWGQ) activation functions.}
\label{fig:activ curves}
\end{figure}

\subsection{Low-bit Activation Quantization Results}

We next compared the performance achieved by adding the $sign$ and
HWGQ $Q(x)$ (backward vanilla ReLU) quantizers to the set-up of the previous section. The results of AlexNet, ResNet-18 and VGG-Variant are summarized in  Table \ref{tab:low precision}. At first, notice that BW is worse than BW+$Q$ of AlexNet in Table \ref{tab:activation comp} due to the impact of the layer re-ordering \cite{DBLP:conf/eccv/RastegariORF16} introduced in Section \ref{subsec:details}. Next, comparing BW to FW+$Q$, where the former binarizes weights only and the
latter quantizes activations only, we observed that weight binarization
causes a minor degradation of accuracy. This is consistent with the
findings of \cite{DBLP:conf/eccv/RastegariORF16,DBLP:journals/corr/HubaraCSEB16}. On the other hand, activation quantization leads to a nontrivial loss.
This suggests that the latter is a more difficult problem than the
former. This observation is not surprising since, unlike weight
binarization, the learning of an activation-quantized
networks needs to propagate gradients through every nondifferentiable
quantization layer.

When weight binarization and activation quantization were combined,
recognition performance dropped even further. For AlexNet, the
drop was much more drastic for BW+$sign$ (backward hard tanh) than for
BW+$Q$ (backward vanilla ReLU). These results support the hypotheses of
Section \ref{sec:binary networks} and \ref{sec:hwgq}, as well
as the findings of Table \ref{tab:activation comp}. The training errors
of BW+$sign$ and BW+$Q$ of AlexNet are shown in Figure \ref{fig:activ curves}. Note the much lower training error of $Q(x)$, suggesting that it
enables a much better approximation of the full precision activations
than $sign(x)$. Nevertheless, the gradient mismatch due to the use
of $Q(x)$ as forward and the vanilla ReLU as backward approximators
made the optimization somewhat instable. For example, the error
curve of BW+$Q$ is bumpy during training. This problem becomes more serious
for deeper networks. In fact, for the ResNet-18 and VGG-Variant, BW+$Q$
performed worse than BW+$sign$. This can be explained by the fact that
the $sign$ has a smaller gradient mismatch problem than the vanilla ReLU.
As will be shown in the following section, substantial improvements are
possible by correcting the mismatch between the forward quantizer $Q(x)$
and its backward approximator.

\begin{table}[t]
\centering \scriptsize \setlength{\tabcolsep}{5.0pt}
\vspace{0.1cm} \caption{Backward Approximations Comparison.}
\label{tab:optimization}
\begin{tabular}
{c|c||c|c|c|c|c}\hline
\multicolumn{2}{c||}{Model} &BW &no-opt &vanilla &clipped &log-tailed\\\hline\hline
\multirow{2}{*}{AlexNet} &Top-1 &52.4  &30.0  &46.8 &48.6  &49.0\\
&Top-5   &75.9  &53.6  &71.0 &72.8  &73.1\\\hline
\multirow{2}{*}{ResNet-18} &Top-1  &61.3  &34.2  &33.0 &54.5  &53.5\\
&Top-5 &83.6  &59.6  &56.9 &78.5  &77.7\\\hline
\multirow{2}{*}{VGG-Variant} &Top-1   &65.5  &42.8  &44.1 &60.9  &60.6\\
&Top-5   &86.5  &68.3  &68.7 &83.2  &82.9\\\hline
\end{tabular}
\end{table}

\begin{figure*}[!t]
\begin{minipage}[b]{.33\linewidth}
\centering
\centerline{\epsfig{figure=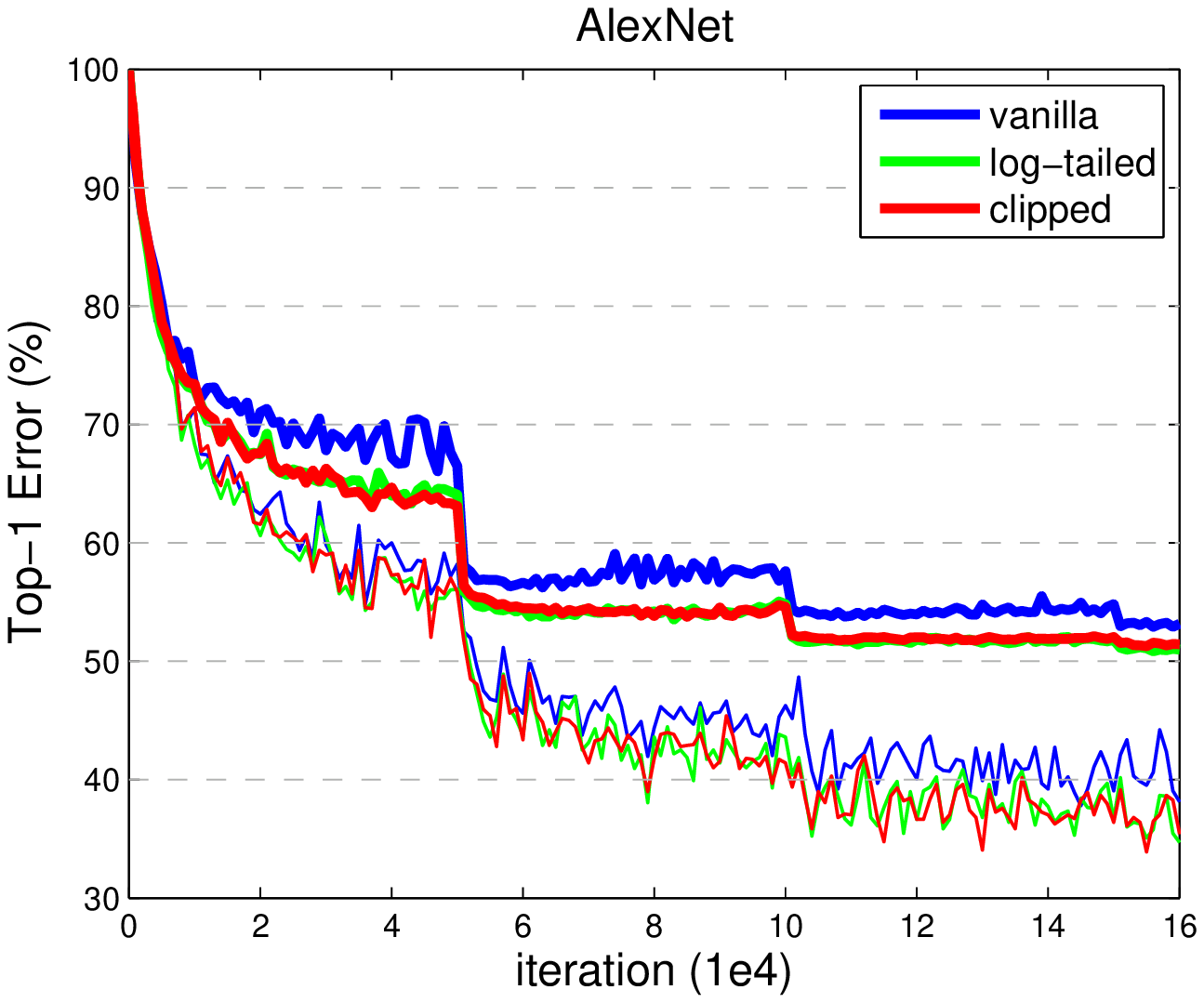,width=6cm,height=4.5cm}}
\end{minipage}
\hfill
\begin{minipage}[b]{.33\linewidth}
\centering
\centerline{\epsfig{figure=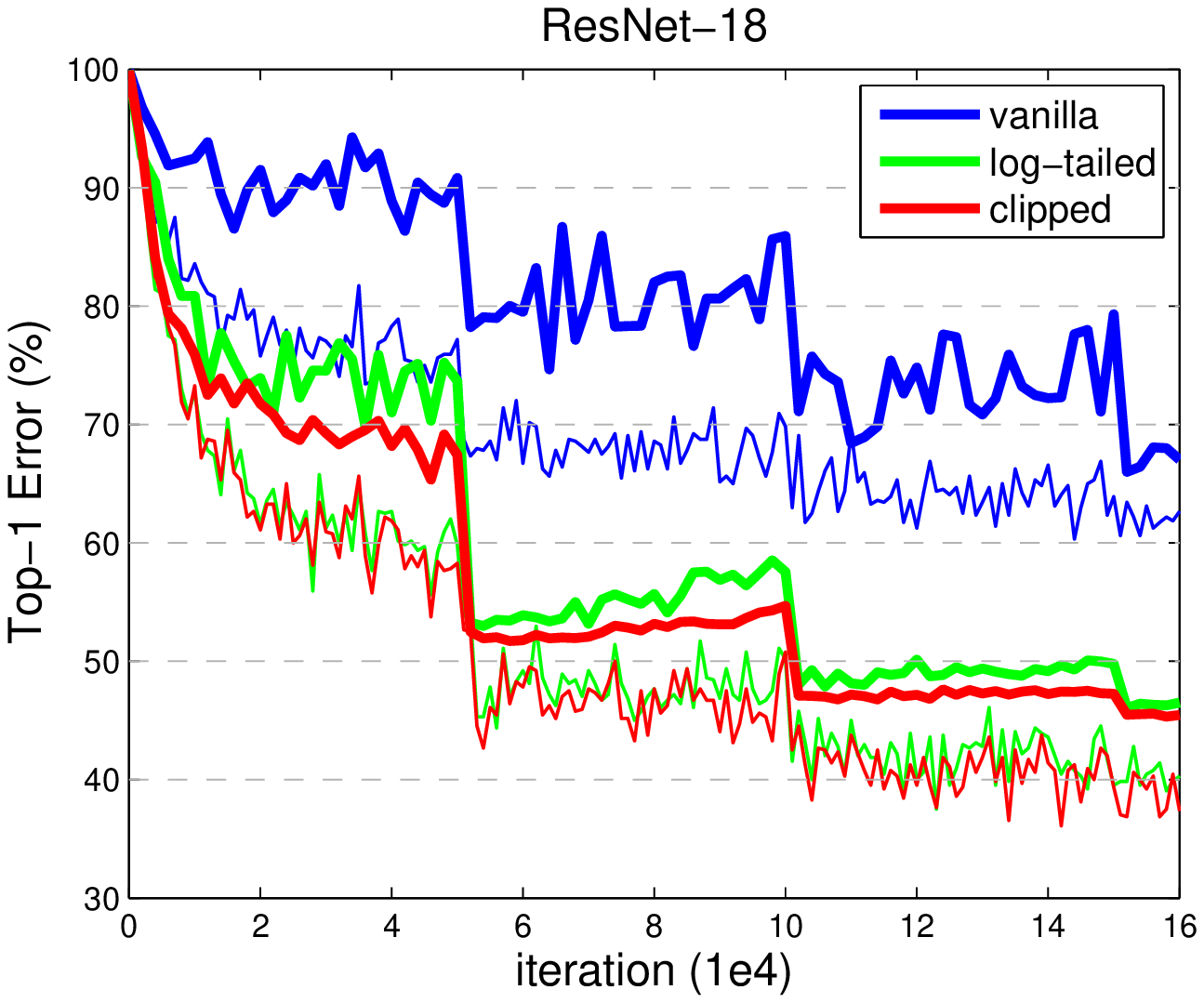,width=6cm,height=4.5cm}}
\end{minipage}
\hfill
\begin{minipage}[b]{.33\linewidth}
\centering
\centerline{\epsfig{figure=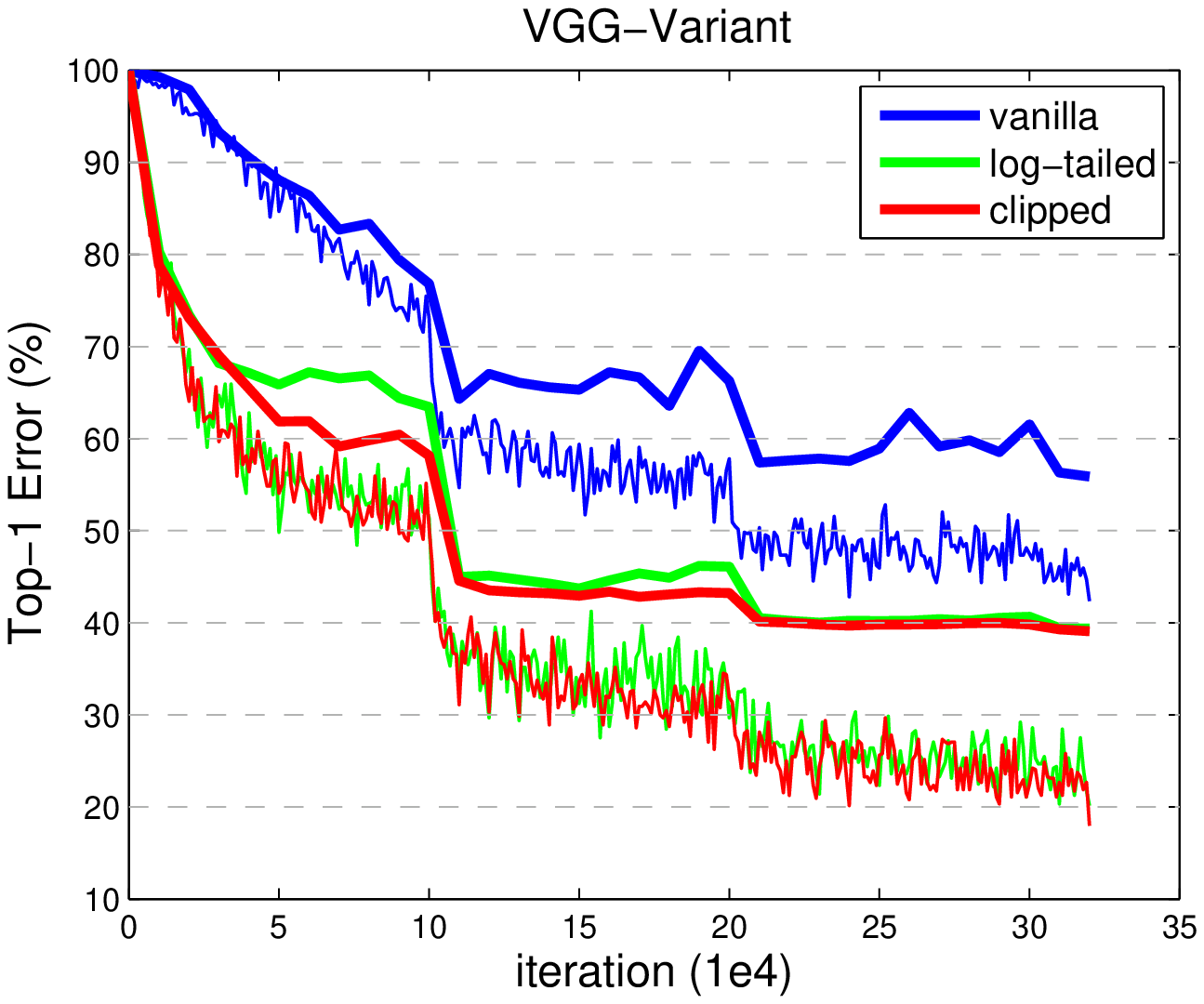,width=6cm,height=4.5cm}}
\end{minipage}
\caption{The error curves of training (thin) and test (thick) for alternative backward approximations.}
\label{fig:relu comp}
\end{figure*}

\subsection{Backward Approximations Comparison}

We next considered the impact of the backward approximators of
Section \ref{subsec:backward relus}.  Table \ref{tab:optimization} shows
the performance achieved by the  three networks under the different
approximations. In all cases, weights were binarized and the HWGQ was used as forward approximator (quantizer). ``no-opt''
refers to the quantization of activations of pre-trained BW networks.
This requires no nondifferentiable approximation, but fails to account for
the quantization error. We attempted to minimize the impact
of cumulative errors across the network by recomputing
the means and variances of all batch normalization layers. Even after
this, ``no-opt'' had significantly lower accuracy than the full-precision
activation networks.

Substantial gains were obtained by training the
activation quantized networks from scratch. Although the vanilla ReLU
had reasonable performance as backwards approximator for AlexNet, much better
results were achieved with the clipped ReLU of~(\ref{equ:clipped relu}) and
the log-tailed ReLU of~(\ref{equ:log relu}). Figure \ref{fig:relu comp} shows
that the larger gradient mismatch of the vanilla ReLU
created instabilities in the optimization, for all networks.
However, these instabilities were more serious for the deeper
networks, such as ResNet-18 and VGG-Variant. This explains
the sharper drop in performance of the vanilla ReLU for these
networks, in Table~\ref{tab:optimization}. Note, in
Figure \ref{fig:relu comp}, that the clipped ReLU and the log-tailed ReLU
enabled more stable learning and reached a much better optimum for all
networks. Among them, the log-tailed ReLU performed slightly better than
the clipped ReLU on AlexNet, but slightly worse on ResNet-18 and VGG-Variant. To be consistent, ``clipped ReLU'' was used in the following sections.

\begin{table}[t]
\centering \scriptsize \setlength{\tabcolsep}{5.0pt}
\vspace{0.1cm} \caption{Bit-width Comparison of Activation Quantization.}
\label{tab:bitwidth}
\begin{tabular}
{c|c||c|c|c|c||c|c||c}\hline
\multicolumn{2}{c||}{quantization type}
&\multicolumn{4}{|c||}{non-uniform} &\multicolumn{2}{|c||}{uniform} &none\\\hline
\multicolumn{2}{c||}{$\#$ levels} &2 &3 &7 &15 &3$^*$ &7$^*$ &BW\\\hline\hline
\multirow{2}{*}{AlexNet} &Top-1 &48.6  &50.6  &52.4 &52.6 &50.5 &51.9 &52.4\\
&Top-5 &72.8  &74.3  &75.8 &76.2 &74.6 &75.7 &75.9\\\hline
\multirow{2}{*}{ResNet-18} &Top-1  &54.5  &57.6 &60.3 &60.8 &56.1 &59.6 &61.3\\
&Top-5  &78.5  &81.0  &82.8 &83.4 &79.7 &82.4 &83.6\\\hline
\end{tabular}
\end{table}

\subsection{Bit-width Impact}

The next set of experiments studied the bit-width impact of the activation quantization. In all cases, weights were binarized. Table \ref{tab:bitwidth} summarizes the
performance of AlexNet and ResNet-18 as a function of the number
of quantization levels. While the former improved with the latter,
there was a saturation effect. The default  HWGQ configuration, also used in all previous experiments, consisted of two non-uniform
positive quantization levels plus a ``0''. This is denoted as ``2'' in the
table. For AlexNet, this very low-bit quantization sufficed to achieve
recognition rates close to those of the full-precision activations.
For this network, quantization with seven non-uniform levels was sufficient
to reproduce the performance of full-precision activations.
For ResNet-18, however, there was a more noticeable gap between low-bit
and full-precision activations. For example, ``3''
outperformed ``2'' by 3.1 points for ResNet-18, but only 2.1
points for AlexNet. These results suggest that increasing the number of
quantization levels is more beneficial for ResNet-18 than for AlexNet.

So far we have used non-uniform quantization.
As discussed in Section \ref{subsec:gaussian}, this requires more bits than uniform quantization during arithmetic computation. Table \ref{tab:bitwidth} also shows the results obtained with uniform
quantization, with superscript ``$*$''. Interestingly, for the same number of quantization levels,
the performance of the uniform quantizer was only slightly worse than
that of its non-uniform counterpart. But for the same bit width, the uniform quantizer was noticeably superior than the non-uniform quantizer, by comparing ``2'' and ``3$^*$'' (both of them need 2-bit representation during arithmetic computation).

\begin{table}[t]
\centering \scriptsize \setlength{\tabcolsep}{5.0pt}
\vspace{0.1cm} \caption{The results of various popular networks.}
\label{tab:final networks}
\begin{tabular}
{c|c||c|c|c}\hline
\multicolumn{2}{c||}{Model} &Reference &Full &HWGQ\\\hline\hline
\multirow{2}{*}{AlexNet} &Top-1 &57.1 &58.5  &52.7\\
&Top-5   &80.2 &81.5  &76.3\\\hline
\multirow{2}{*}{ResNet-18} &Top-1  &69.6 &67.3  &59.6\\
&Top-5 &89.2 &87.9  &82.2\\\hline
\multirow{2}{*}{ResNet-34} &Top-1  &73.3 &69.4  &64.3\\
&Top-5 &91.3 &89.1  &85.7\\\hline
\multirow{2}{*}{ResNet-50} &Top-1  &76.0 &71.5  &64.6\\
&Top-5 &93.0 &90.5  &85.9\\\hline
\multirow{2}{*}{VGG-Variant} &Top-1  &- &69.8  &64.1\\
&Top-5  &- &89.3  &85.6\\\hline
\multirow{2}{*}{GoogLeNet} &Top-1  &68.7 &71.4  &63.0\\
&Top-5 &88.9 &90.5  &84.9\\\hline
\end{tabular}
\end{table}

\begin{table}[t]
\centering \scriptsize \setlength{\tabcolsep}{5.0pt}
\vspace{0.1cm} \caption{Comparison with the state-of-the-art low-precision methods. Top-1 gap to the corresponding full-precision networks is also reported.}
\label{tab:state-of-the-art}
\begin{tabular}
{c||c|c|c||c|c}\hline
\multirow{2}{*}{Model}
&\multicolumn{3}{|c||}{AlexNet}
&\multicolumn{2}{|c}{ResNet-18}\\
\cline{2-6}
&XNOR &DOREFA & HWGQ & XNOR & HWGQ\\\hline
Top-1 &44.2 &47.7 &52.7 &51.2 &59.6\\
Top-5 &69.2 &- &76.3 &73.2 &82.2\\\hline
Top-1 gap &-12.4 &-8.2 &-5.8 &-18.1 &-7.7\\\hline
\end{tabular}
\end{table}

\subsection{Comparison with the state-of-the-art}
\label{subsec:state-of-the-art}

Table \ref{tab:final networks}\footnote{The reference performance of AlexNet and GoogLeNet is at https://github.com/BVLC/caffe, and of ResNet is at https://github.com/facebook/fb.resnet.torch. Our worse ResNet implementations are probably due to fewer training iterations and no further data augmentation.} presents a comparison between the full
precision and the low-precision HWGQ-Net of many popular network
architectures. For completeness, we
consider the GoogLeNet, ResNet-34 and ResNet-50 in this section. In all
cases, the HWGQ-Net used 1-bit binary weights, a 2-bit uniform HWGQ as forward approximator, and the clipped ReLU as backwards approximator.
Comparing to the previous ablation experiments, the numbers of training
iterations were doubled and polynomial learning rate annealing
(power of 1) was used for HWGQ-Net, where it gave a
slight improvement over step-wise annealing. Table \ref{tab:final networks}
shows that the HWGQ-Net approximates well all popular networks,
independently of their complexity or depth. The top-1 accuracy drops from full- to low-precision are similar for all networks (5$\sim$9 points),
suggesting that low-precision HWGQ-Net will achieve improved performance
as better full-precision networks become available.

Training a network with binary weights and low-precision activations from
scratch is a new and challenging problem, only addressed by a few
previous works \cite{DBLP:journals/corr/HubaraCSEB16,DBLP:conf/eccv/RastegariORF16,DBLP:journals/corr/ZhouNZWWZ16}.
Table \ref{tab:state-of-the-art} compares the HWGQ-Net with the recent XNOR-Net \cite{DBLP:conf/eccv/RastegariORF16} and DOREFA-Net
\cite{DBLP:journals/corr/ZhouNZWWZ16}, on the ImageNet classification task.
The DOREFA-Net result is for a model of binary weights, 2-bit activation,
full precision gradient and no pre-training. For AlexNet, the
HWGQ-Net outperformed the XNOR-Net and the DOREFA-Net by a large margin.
Similar improvements over the XNOR-Net were observed for the ResNet-18,
where DOREFA-Net results are not available. It is worth noting that the gaps between the full-precision networks and the HWGQ-Net (-5.8 for AlexNet
and -7.7 for ResNet-18) are much smaller than those of the
XNOR-Net (-12.4 for AlexNet and -18.1 for ResNet-18) and the DOREFA-Net (-8.2 for AlexNet). This is strong evidence that the HWGQ
is a better activation quantizer. Note that, in contrast to the
experimentation with one or two networks by
\cite{DBLP:journals/corr/HubaraCSEB16,DBLP:conf/eccv/RastegariORF16,DBLP:journals/corr/ZhouNZWWZ16},
the HWGQ-Net is shown to perform well for various network architectures. To the best of our knowledge, this is the first time that a single low-precision network is shown to successfully approximate many popular networks.

\subsection{Results on CIFAR-10}

In addition, we conducted some experiments on the CIFAR-10
dataset \cite{krizhevsky2009learning}. The network structure used, denoted
VGG-Small, was similar to that of \cite{DBLP:journals/corr/HubaraCSEB16} but relied on a cross-entropy loss and without two fully connected layers. The learning strategy was that used in the VGG-Variant of Section \ref{subsec:details}, but the mini-batch size was 100 and the learning rate was divided by 10 after every 40K iterations (100K in total). The data augmentation strategy of \cite{DBLP:journals/corr/HeZRS15} was used. As in Section \ref{subsec:state-of-the-art}, polynomial learning rate decay was used for low-precision VGG-Small. The HWGQ-Net results are compared with the state-of-the-art in Table \ref{tab:cifar10}. It can be seen that the HWGQ-Net achieved better performance than various popular full-precision networks, e.g. Maxout \cite{DBLP:conf/icml/GoodfellowWMCB13},
NIN \cite{DBLP:journals/corr/LinCY13}, DSN \cite{DBLP:conf/aistats/LeeXGZT15}, FitNet \cite{DBLP:journals/corr/RomeroBKCGB14}, and than various low precision
networks. The low-precision VGG-Small drops 0.67 points when compared to
its full-precision counterpart. These findings are consistent with those
on ImageNet.

\begin{table}[t]
\centering \scriptsize \setlength{\tabcolsep}{12.0pt}
\vspace{0.1cm} \caption{The results on CIFAR-10. The bit width before and after ``+'' is for weights and activations respectively.}
\label{tab:cifar10}
\begin{tabular}
{c|c|c}\hline
precision &{Method} &error (\%)\\\hline
\multirow{6}{*}{Full + Full} &Maxout \cite{DBLP:conf/icml/GoodfellowWMCB13} &9.38\\
&NIN \cite{DBLP:journals/corr/LinCY13}  &8.81\\
&DSN \cite{DBLP:conf/aistats/LeeXGZT15}  &8.22\\
&FitNet \cite{DBLP:journals/corr/RomeroBKCGB14}  &8.39\\
&ResNet-110 \cite{DBLP:journals/corr/HeZRS15}  &6.43\\
&VGG-Small  &6.82\\\hline
1-bit + Full &BinaryConnect \cite{DBLP:conf/nips/CourbariauxBD15} &8.27\\
2-bit + Full &Ternary Weight Network \cite{DBLP:journals/corr/LiL16}   &7.44\\\hline
1-bit + 1-bit &BNN \cite{DBLP:journals/corr/HubaraCSEB16} &10.15\\
1-bit + 2-bit &VGG-Small-HWGQ   &7.49\\\hline
\end{tabular}
\end{table}

\section{Conclusion}

In this work, we considered the problem of training high performance
deep networks with low-precision. This was achieved by
designing two approximators for the ReLU non-linearity.
The first is a half-wave Gaussian quantizer, applicable in the
feedforward network computations. The second is a piece-wise continuous
function, to be used in the backpropagation step during learning.
This design overcomes the learning inefficiency of the popular binary
quantization procedure, which produces a similar approximation for the
less effective hyperbolic tangent nonlinearity. To minimize the
problem of gradient mismatch, we have studied several backwards approximation functions. It was shown that the mismatch is most affected
by activation outliers. Insights from the robust estimation
literature were then used to propose the clipped ReLU and
log tailed ReLU approximators. The network that results from
the combination of these with the HWGQ, denoted HWGQ-Net
was shown to significantly outperform previous efforts at
deep learning with low precision, substantially reducing the gap between
the low-precision and full-precision various state-of-the-art networks. These promising experimental results suggest that the HWGQ-Net can be very useful for the deployment of state-of-the-art neural networks in real world applications.

{\small
\bibliographystyle{ieee}
\bibliography{egbib}
}

\end{document}